\documentclass[10pt,twocolumn,twoside]{IEEEtran}
\IEEEoverridecommandlockouts

\usepackage{ifpdf}
\usepackage[pdftex]{graphicx}
\usepackage{algorithm}
\usepackage{algorithmic}
\usepackage{graphicx}
\usepackage{amsmath}
\usepackage{amssymb}
\usepackage{subfigure}
\usepackage{epstopdf}
\usepackage{multirow}
\usepackage{array}
\usepackage{tikz}
\usepackage{color}
\usepackage{stfloats}
\usepackage{amsmath,amsthm}
\ifCLASSINFOpdf
\else
\fi

\begin{document}
%
\title{Locally-Supervised Deep Hybrid Model for Scene Recognition}
%
%
%

\author{Sheng~Guo,
        Weilin~Huang,~\IEEEmembership{Member,~IEEE,}
        Limin~Wang,
        and~Yu~Qiao,~\IEEEmembership{Senior~Member,~IEEE}
\thanks{This work is partly supported by National High-Tech Research and Development Program of China $($2016YFC1400704$)$, National Natural Science Foundation of China (61503367), Guangdong Research Program (2014B050505017,2015B010129013,2015A030310289), External Cooperation Program of BIC Chinese Academy of Sciences (172644KYSB20150019), and Shenzhen Research Program (JSGG20150925164740726, JCYJ20150925 163005055, CXZZ20150930104115529).}
\thanks{S. Guo is with Shenzhen College of Advanced Technology, University of Chinese Academy of Sciences,
and is with Shenzhen Institutes of Advanced Technology, Chinese Academy of Sciences, Shenzhen, China (e-mail:guosheng1001@gmail.com)}
\thanks{W. Huang is with Shenzhen Institutes of Advanced Technology, Chinese Academy of Sciences, Shenzhen, China, (e-mail: wl.huang@siat.ac.cn)}
\thanks{L. Wang was with Shenzhen Institutes of Advanced Technology, Chinese Academy of Sciences, Shenzhen, China, and is with Computer Vision Laboratory, ETH Zurich, Switzerland, (e-mail: 07wanglimin@gmail.com) }
\thanks{Y. Qiao is with Shenzhen Key Lab of Computer Vision and Pattern Recognition, Shenzhen Institutes of Advanced Technology, Chinese Academy of Sciences, Shenzhen, China, and is with Department of Information Engineering, The Chinese University of Hong Kong (e-mail: yu.qiao@siat.ac.cn)}}
\maketitle

\begin{abstract}
Convolutional neural networks (CNN) have recently achieved remarkable successes in various image classification and understanding tasks. The deep features obtained at the top fully-connected layer of the CNN (FC-features) exhibit rich global semantic information and are extremely effective in image classification. On the other hand, the convolutional features in the middle layers of the CNN also contain meaningful local information, but are not fully explored for image representation. In this paper, we propose a novel Locally-Supervised Deep Hybrid Model (LS-DHM) that effectively enhances and explores the  convolutional features for scene recognition.
Firstly, we notice that the  convolutional features capture local objects and fine structures of scene images, which yield important cues for discriminating ambiguous scenes, whereas these features are significantly eliminated in the highly-compressed FC representation.
  Secondly, we propose a new Local Convolutional Supervision (LCS) layer to enhance the local structure of the image by directly propagating the label information to the convolutional layers.
  Thirdly, we propose an efficient Fisher Convolutional Vector (FCV) that successfully rescues the orderless mid-level semantic information (e.g. objects and textures) of scene image. The FCV encodes the large-sized convolutional maps into a fixed-length mid-level representation, and is demonstrated to be strongly complementary to the high-level FC-features.
  Finally, both the FCV and FC-features are collaboratively employed in the LS-DHM representation, which
  achieves outstanding  performance in our experiments. It obtains $83.75\%$ and $67.56\%$ accuracies respectively on the heavily benchmarked MIT Indoor67 and SUN397 datasets, advancing the stat-of-the-art substantially.


\end{abstract}


%
\IEEEpeerreviewmaketitle

\section{Introduction}
\begin{figure}
\centering
\footnotesize
\renewcommand{\arraystretch}{1.3}
\begin{minipage}[l]{1.2\linewidth}

\includegraphics[scale=0.48]{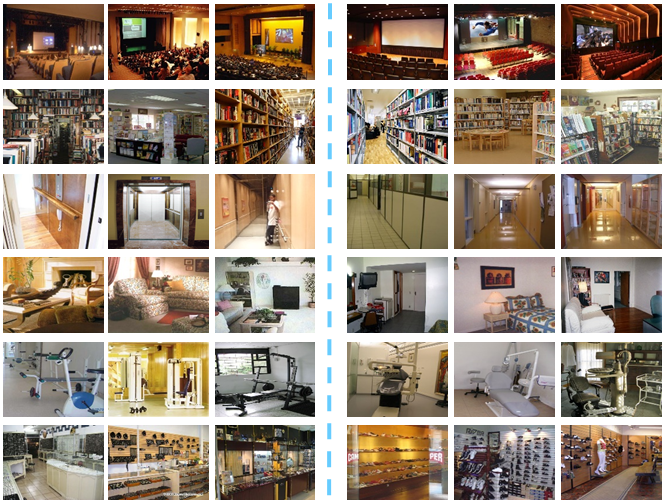}
\end{minipage}

\vskip +0.2cm

\begin{minipage}[l]{1.1\linewidth}
\begin{tabular}{l|l||l|l|l}
\hline
category (left) & category (right) & FC-Fea. & Conv.-Fea. & Both\\\hline
$auditorium$ & $movietheater$ & 38.9&22.2& 11.1 \\
$bookstore$ & $library$ & 25.0 & 10.0 & 5.0 \\
$elevator$ & $corridor$ & 9.5&4.8 & 4.8\\
$livingroom$ & $bedroom$ & 15.0& 10.0 & 0.0 \\
$gym$ & $dental office$ & 11.1 &5.6& 0.0 \\
$jewelleryshop$ & $shoeshop$ & 9.1&4.6 & 0.0\\
\hline
\end{tabular}
\end{minipage}

\caption{\textbf{Top Figure}: category pairs with similar global layouts, which are difficult to be discriminated by purely using high-level fully-connected features (FC-features). The category names are listed in the bottom table. \textbf{Bottom Table}: classification errors (\%) between paired categories by using the convolutional features, FC-features, or both of them. }
\label{fig3}
\end{figure}

     \IEEEPARstart {H}{uman} has a remarkable ability to categorize complex scenes very accurately and rapidly. This ability is important for human to infer the current situations and navigate the environments \cite{oliva2005gist}. Computer scene recognition and understanding aims at imitating this human ability by using algorithms to analyze input images. This is a fundamental problem in computer vision, and plays a crucial role on the success of numerous application areas like image retrieval, human machine interaction, autonomous driving, etc.

    The difficulties of scene recognition come from several aspects. Firstly, scene categories are defined not only by various image contents they contain, such as local objects and background environments, but also by global arrangements, interactions or actions between them, such as eating in restaurants, reading in library, watching in cinema. These cause a large diversity of the scene contents which imposes a huge number of scene categories and large within-class variations. These make it much more challenging than the task of object classification. Furthermore, scene images often include numerous fine-grained categories which exhibit very similar contents and structures, as shown in Fig. \ref{fig3}. These fine-grained categories are hard to be discriminated by purely using the high-level FC-features of CNN, which often capture highly abstractive and global layout information. These difficulties make it challenging to develop a robust yet discriminative method that accounts for all types of feature cues for scene recognition.


    Deep learning models, i.e. CNN \cite{LeCun1998,LeCun1989}, have been introduced for scene representation and classification, due to their great successes in various related vision tasks \cite{LeCun2015,krizhevsky2012imagenet,szegedy2014going,zhou2014learning,Girshick2014,Long2015,Huang2014,He2016,tian2016detecting}. Different from previous methods \cite{doersch2013mid,wang2014latent,juneja2013blocks,guo2014f,sanchez2013image,xie2014orientational,Guo2015,wang2015mofap} that compute hand-crafted features or descriptors, the CNN directly learns high-level features from raw data with multi-layer hierarchical transformations. Extensive researches demonstrate that, with large-scale training data (such as ImageNet \cite{russakovsky2014imagenet,deng2009imagenet}), the CNN can learn effective high-level features at top fully-connected (FC) layer. The FC-features generalize well for various different tasks, such as object recognition \cite{krizhevsky2012imagenet,szegedy2014going,simonyan2014very}, detection \cite{Girshick2014,Girshick2015} and segmentation \cite{Long2015,Hariharan2014}.

    However, it has been shown that directly applying the CNNs trained with the ImageNet \cite{donahue2013decaf} for scene classification was difficult to yield a better result than the leading hand-designed features incorporating with a sophisticated classifier \cite{sanchez2013image}. This can be ascribed to the fact that the ImageNet data \cite{russakovsky2014imagenet} is mainly made up of images containing large-scale objects, making the learned CNN features globally object-centric. To overcome this problem, Zhou \emph{et al.} trained a scene-centric CNN by using a large newly-collected  scene dataset, called Places, resulting in a significant performance improvement \cite{zhou2014learning}. In spite of using different training data, the insight is that the scene-centric CNN is capable of learning more meaningful local structures of the images (e.g. fine-scale objects and local semantic regions) in the convolutional layers, which are crucial to discriminate the ambiguous scenes \cite{zhou2014object}. Similar observation was also presented in  \cite{zeiler2014visualizing} that the neurons at middle convolutional layers exhibit strong semantic information. Although it has been demonstrated that the convolutional features include the important scene cues, the classification was still built on the FC-features in these works, without directly exploring the mid-level features from the convolutional layers \cite{zhou2014learning,wang2015places205}.


    In CNN, the convolutional features are highly compressed when they are forwarded to the FC layer, due to computational requirement (i.e. the high-dimensional FC layer will lead to huge weight parameters and computational cost). For example, in the celebrated AlexNet \cite{krizhevsky2012imagenet}, the $4^{th}$ and $5^{th}$ convolutional layer have 64,896 and 43,264 nodes respectively, which are reduced considerably to 4,096 (about 1/16 or 1/10) in the $6^{th}$ FC layer. And this compression is simply achieved by pooling and transformations with sigmod or ReLU operations. Thus there is a natural question: \textit{ are the fine sematic features learned in the convolutional layers well preserved in the fully-connected layers? } If not, \textit{how to rescue the important mid-level  convolutional features lost when forwarded to the FC layers}. In this paper, we explore the questions in the context of scene classification.

     Building on these observations and insightful analysis, this paper strives for a further step by presenting an efficient approach that both enhances and encodes the local semantic features in the convolutional layers of the CNN. We propose a novel Locally-Supervised Deep Hybrid Model (LS-DHM) for scene recognition, making the following contributions.

     Firstly, we propose a new local convolutional supervision (LCS) layer  built upon the convolutional layers. The LCS layer directly propagates the label information to the low/mid-level convolutional layers, in an effort to enhance the mid-level semantic information existing in these layers. This avoids the important scene cues to be undermined by transforming them through the highly-compressed FC layers.

     Secondly, we develop the Fisher Convolutional Vector (FCV) that effectively encodes meaningful local detailed information by pooling the convolutional features into a fixed-length representation. The FCV rescues  rich semantic information of local fine-scale objects and regions by extracting mid-level features from the convolutional layers, which endows it with strong ability to discriminate the ambiguous scenes. At the same time, the FCV discards explicit spatial arrangement by using the FV encoding, making it robust to various local image distortions.

     Thirdly, both the FCV and the FC-features are collaboratively explored in the proposed LS-DHM representation. We demonstrate that the FCV with LCS enhancement is strongly complementary to the high-level FC-features, leading to significant performance improvements. The LS-DHM achieves 83.75\% and 67.56\% accuracies on the MIT Indoor67 \cite{quattoni2009recognizing} and SUN397 \cite{xiao2010sun}, remarkably outperforming all previous methods.

     The rest of paper is organized as follows. Related studies are briefly reviewed in Section II. Then the proposed Locally-Supervised Deep Hybrid Model (LS-DHM), including the local convolutional supervision (LCS) layer and the Fisher Convolutional Vector (FCV), is described in Section III. Experimental results are compared and discussed in Section IV, followed by the conclusions in Section V.

\section{Related Works}

Scene categorization is an important task in computer vision and image related applications. Early methods utilized hand-crafted holistic features, such as GIST \cite{oliva2005gist}, for scene representation. Holistic features are usually computationally efficient but fail to deliver rich semantic information, leading to poor performance for indoor scenes with man-made objects \cite{wu2011centrist}. Later Bag of Visual Words (e.g. SIFT \cite{lowe2004distinctive}, HoG \cite{dalal2005histograms}) and its variants (e.g. Fisher vector \cite{sanchez2013image}, Sparse coding \cite{lazebnik2006beyond})  became popular in this research area. These methods extract dense local descriptors from input image, then encode and pool these descriptors into a fixed length representation for classification. This representation contains abundant statistics of local regions and achieves good performance in practice. However, local descriptors only exhibit limited semantic meaning and global spatial relationship of local descriptors is generally ignored in these methods. To relieve this problem, semantic part based methods are proposed. Spatial Pyramid Matching (SPM) \cite{lazebnik2006beyond}, Object Bank (OB) \cite{li2010object} and Deformable Part based Model (DPM) \cite{pandey2011scene} are examples along this line.

However, most of these approaches used hand-crafted features, which are difficult to be adaptive for different image datasets.  Recently, a number of learning based methods have been developed for image representation. In \cite{shao2014feature}, an evolutionary learning approach was proposed. This methodology automatically generated domain-adaptive global descriptors for image/scene classification, by using multi-objective genetic programming. It can simultaneously extract and fuse the features from various color and gray scale spcaces. Fan and Lin \cite{zhu2014weakly} designed a new visual categorization framework by using a weekly-supervised cross-domain dictionary learning algorithm, with considerable performance imporvements achieved. Zhang \emph{et al.} \cite{zhang2014learning} proposed an Object-to-Class (O2C) distance for scene classification by exploring the Object Bank representation. Based on the O2C distance, they built a kernelization framework that maps the Object Bank representation into a new distance space, leading to a stronger discriminative ability.

 In recent years, CNNs have achieved record-breaking results on standard image datasets, and there have been a number of attempts to develop deep networks for scene recognition \cite{donahue2013decaf,zhou2014learning,wang2016weakly,wang2016knowledge}. Krizhevsky \emph{et al.}  \cite{krizhevsky2012imagenet} proposed a seven-layer CNN, named as AlexNet, which achieved significantly better accuracy than other non-deep learning methods in ImageNet LSVRC 2012.
  Along this direction, two very deep convolutional networks, the GoogleNet \cite{szegedy2014going} and VGGNet \cite{simonyan2014very}, were developed, and they achieved the state-of-the-art performance in LSVRC 2014.
  However, the classical CNNs trained with ImageNet are object-centric which cannot obtain better performance on scene classification than handcrafted features \cite{donahue2013decaf}.
   Recently, Zhou \emph{et al.} developed a scene-centric dataset called Places, and utilized it to train the CNNs, with significantly performance improvement on scene classification \cite{zhou2014learning}. Gong \emph{et al.} employed Vector of Locally Aggregated Descriptors (VLAD) \cite{jegou2010aggregating} for pooling multi-scale orderless FC-features (MOP-CNN) for scene classification \cite{gong2014multi}. Despite having powerful capabilities, these successful models are all built on the FC representation for image classification.


 \begin{figure}
\centering
\includegraphics[height=2.4in,width=3.4in,angle=0]{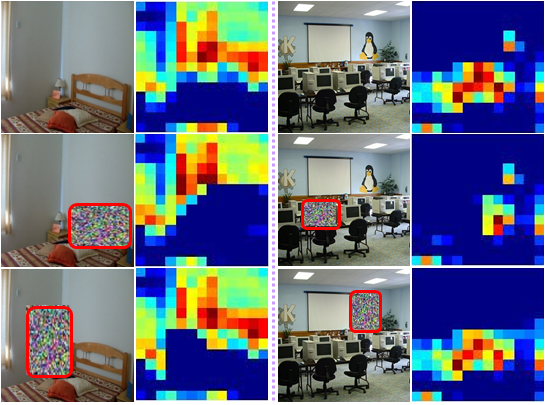}
\caption{\textbf{Top}: images of bedroom  (left) and computer room (right), and  their corresponding convolutional feature maps. \textbf{Middle}: image with key objects occluded, i.e., \emph{bed} or \emph{computers}. \textbf{Bottom}: image with unimportant areas occluded. Occluding key  objects significantly modifies the structures of convolutional maps, while unimportant regions change the convolutional features slightly. This indicates that the convolutional features are crucial to discriminate the key objects in the scene images.}
\label{fig2}
\end{figure}

    The GoogleNet introduces several auxiliary supervised layers which were selectively connected to the middle level convolutional layers \cite{szegedy2014going}. This design encourages the low/mid-level convolutional features to be learned from the label information, avoiding gradient information vanished in the very deep layers.
    Similarly, Lee \emph{et al.} \cite{lee2014deeply} proposed deeply supervised networks (DSN) by adding a auxiliary supervised layer onto each convolutional layer. Wang \emph{et al.} employed related methods for scene recognition by selectively adding the auxiliary supervision into several convolutional layers \cite{wang2015training}.
    Our LCS layer is motivated from these approaches, but it has obvious distinctions by design. The final label is directly connected to the convolutional layer of the LCS, allowing the label to directly supervise each activation in the convolutional layers, while all related approaches keep the FC layers for connecting the label and last convolutional layer \cite{szegedy2014going,lee2014deeply,wang2015training}.
     Importantly, all these methods use the FC-features for classification, while our studies focus on exploring the convolutional features enhanced by the LCS.

     Our work is also related to several recent efforts on exploring the convolutional features for object detection and classification. Oquab \emph{et al.} \cite{oquab2014learning} demonstrated that the rich mid-level features of CNN pre-trained on the large ImageNet data can been applied to a different task, such as object or action recognition and  localization. Sermanet \emph{et al.} explored Sparse Coding to encode the convolutional and FC features for pedestrian detection \cite{sermanet2013pedestrian}. Raiko \emph{et al.} transformed the outputs of each hidden neuron to have zero output and slope on average, making the model advanced in training speed and also generalized better \cite{raiko2012deep}. Recently, Yang and Ramanan \cite{Yang2015} proposed directed acyclic graph CNN (DAG-CNN) by leveraging multi-layer convolutional features for scene recognition. In this work, the simple average pooling was used for encoding the convolutional features. Our method differs from these approaches by designing a new LCS layer for local enhancement, and developing the FCV for features encoding with the Fisher kernel.

    Our method is also closed to Cimpoi \emph{et al.}'s work \cite{cimpoi2015deep}, where a new texture descriptor, FV-CNN, was proposed. Similarly, the FV-CNN applies the Fisher Vector to encode the convolutional features, and achieves excellent performance on texture recognition and segmentation. However, our model is different from the FV-CNN in CNN model design, feature encoding and application tasks. First, the proposed LCS layer allows our model to be trained for learning stronger local semantic features, immediately setting us apart from the FV-CNN which directly computes the convolutional features from the ``off-the-shelf" CNNs. Second, our LS-DHM uses both the FCV and FC-features, where the FCV is just computed at a single scale, while the FV-CNN purely computes multi-scale convolutional features for image representation, e.g. ten scales. This imposes a significantly larger computational cost, e.g. about 9.3 times of our FCV. Third, the application tasks are different. The FV-CNN is mainly developed for texture  recognition, where the global spatial layout is not crucial, so that the FC-features are not explored. In contrast, our scene recognition requires both global  and local fine-scale information, and our LS-DHM allows both FCV and FC-features to work collaboratively, which eventually boost the performance.


\section{Locally-Supervised Deep Hybrid Model}

%


 In this section, we first  discuss and analyze the properties of convolutional features of the CNN networks. In particular, we pay special attention on the difference of scene semantics computed by the convolutional layers and the FC layers. Then we present details of the proposed Locally-Supervised Deep Hybrid Model (LS-DHM) that computes multi-level deep features. It includes a newly-developed local convolutional supervision (LCS) layer to enhance the convolutional features, and utilizes the Fisher Convolutional Vector (FCV) for encoding the convolutional features.
 Finally, we discuss the properties of the LS-DHM by making comparisons with related methods, and explain insights that eventually lead to performance boost.


\subsection{Properties of Convolutional Features}

 \begin{figure*}
\centering
\subfigure[\emph{Bakery} category]{\includegraphics[height=1.4in,width=3.5in,angle=0]{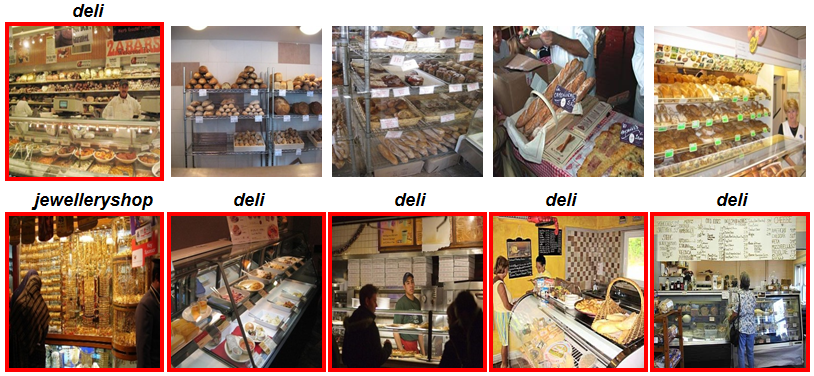}}
\subfigure[\emph{Church-inside} category]{\includegraphics[height=1.4in,width=3.5in,angle=0]{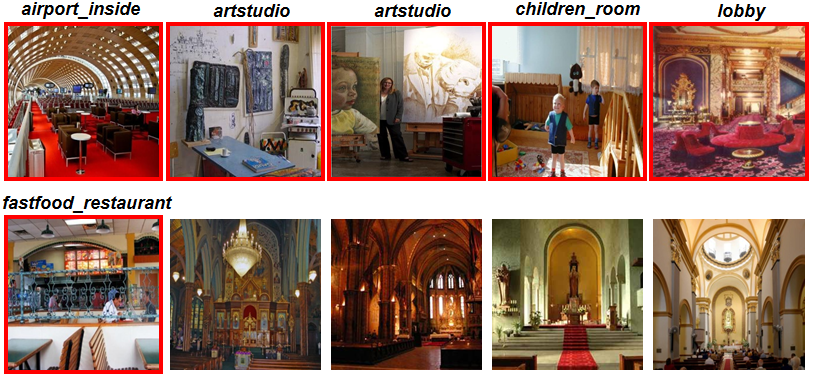}}
\caption{The classification results of the \emph{Bakery} and \emph{Church-inside} categories. We list the images with the \emph{lowest five} classification scores by using the convolutional features (\textbf{top row}) and the FC-features (\textbf{bottom row}). The images with higher scores are generally classified correctly by each type of feature. The image with incorrect classification is labeled by a RED bounding box. We observe that the convolutional features perform better on the \emph{Bakery} category which can be mainly discriminated by the iconic objects, while the FC-features got better results on the \emph{Church-inside} category where the global layout information dominate. The FC-features are difficult to discriminate the \emph{Bakery} and the \emph{Deli}, which have very closed global structures, but are distinctive in local objects contained. These observations inspire our incorporation of both types of features for scene categorization.}
\label{fig:ConvFC}
\end{figure*}

\begin{figure*}
\centering
\subfigure{\includegraphics[height=1.8in,width=3.5in,angle=0]{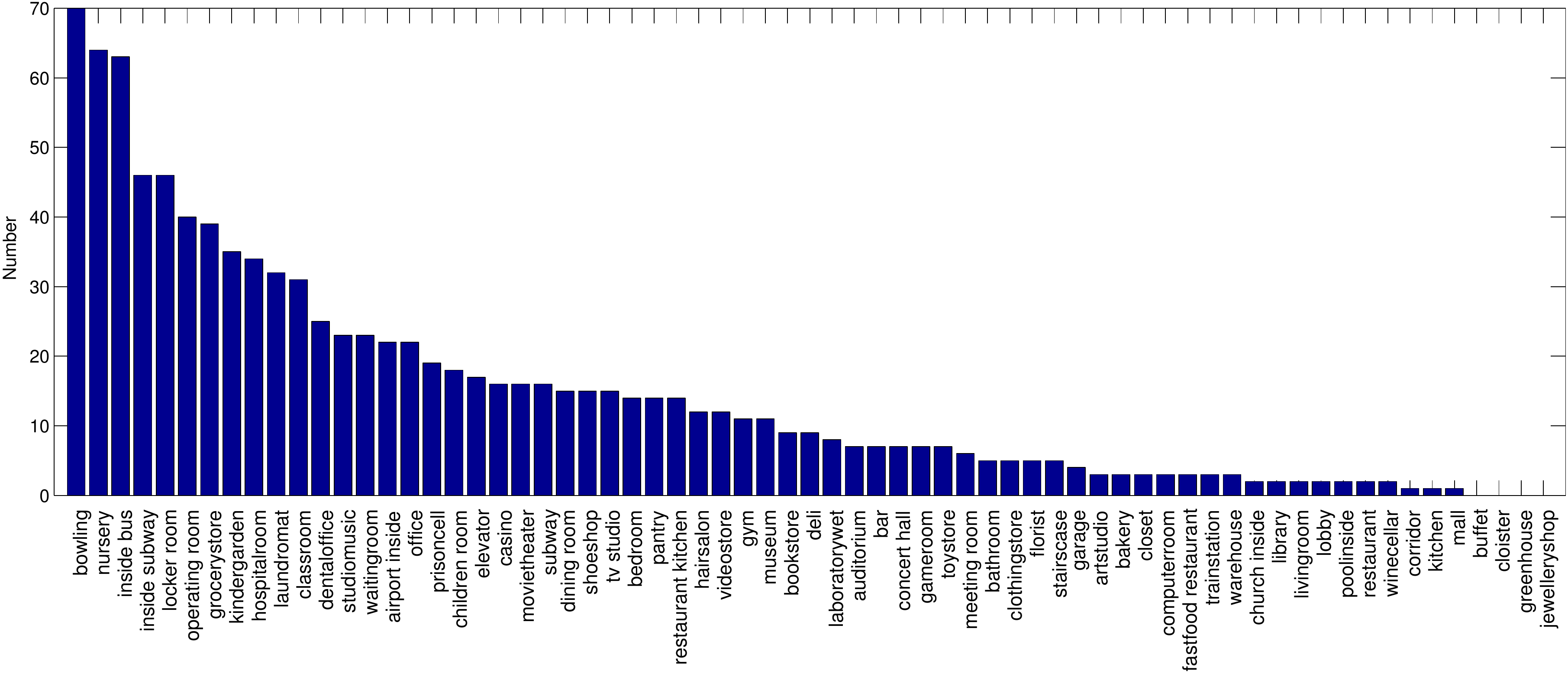}}
\subfigure{\includegraphics[height=1.8in,width=3.5in,angle=0]{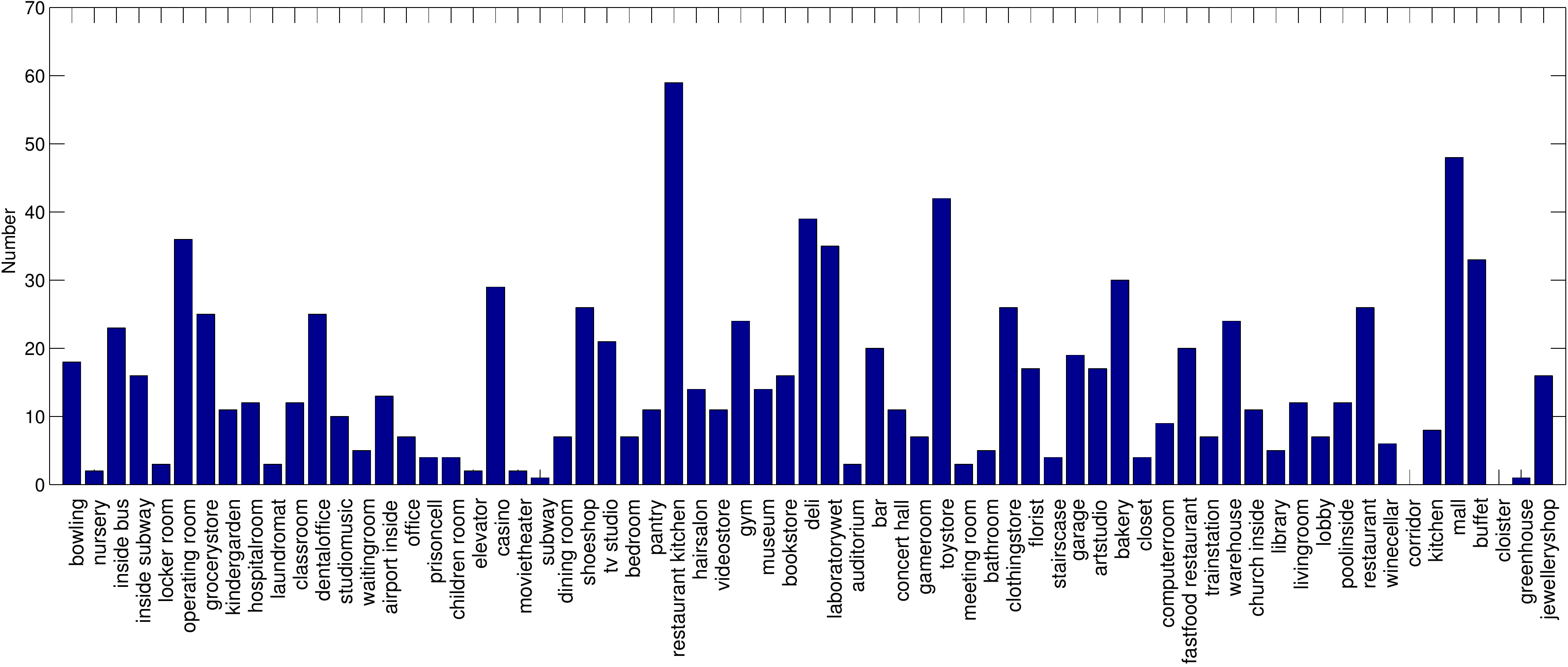}}
\caption{Distributions of top 1,000 images with the largest average activations in the FC layer (left) and the convolutional layer (right).  The  average activation for each image is the average value of all activations in the 7th FC layer or 4th convolutional layer of the AlexNet.}
\label{fig7}
\end{figure*}

  The remarkable success of the CNN encourages researchers to explore the properties of the CNN features, and to understand why they work so well. In \cite{zeiler2014visualizing},  Zeiler and Fergus introduced deconvolutional network to visualize the feature activations in different layers. They shown that the CNN features exhibit increasing invariance and class discrimination as we ascend layers. Yosinski \emph{et al.} \cite{yosinski2014transferable} analyzed the transferability of CNN features learned at various layers, and found the top layers are more specific to the training tasks. More recently, Zhou \emph{et al.} \cite{zhou2014object} show that certain nodes in the Places-CNN, which was trained on the scene data without any object-level label, can surprisingly learn strong object information automatically. Xie \emph{et al.} \cite{xie2016hybrid} propose a hybrid representation method for scene recognition and domain adaptation by integrating the powerful CNN features with the traditional well-studied dictionary-based features. Their results demonstrate that the CNN features in different layers correspond to multiple levels of scene abstractions, such as \emph{edges}, \emph{textures}, \emph{objects}, and \emph{scenes}, from low-level to high-level. A crucial issue is which levels of these abstractions are discriminative yet robust for scene representation.

   Generally, scene categories can be discriminated by their global spatial layouts. This \emph{scene}-level distinctions can be robustly captured by the FC-features of CNN. However, there also exist a large number of ambiguous categories, which do not have distinctive global layout structure. As shown in Fig. \ref{fig3}, it is more accurate to discriminate these categories by the iconic objects within them. For instance, the $bed$ is the key object to identify the $bedroom$, making it crucial to discriminate the $bedroom$ and $living room$. While the $jewelleryshop$ and $shoeshop$ have a similar global layout, the main difference lies in the subtle object information they contain, such as $jewellery$ and $shoe$. Obviously, the key object information provides important cues for discriminating these ambiguous scenes, and the mid-level convolutional features capture rich such object-level and fine structure information. We conduct a simple experiment by manually occluding  a region of the image. As shown in Fig. \ref{fig2}, the convolutional feature maps (from the $4^{th}$ convolutional layer) are affected significantly if the key objects defining the scene categories are occluded ($2^{nd}$ row), while the maps show robustness to the irrelevant objects or regions ($3^{rd}$ row).  These results and discussions suggest that \textit{the middle-level convolutional activations are highly sensitive to the presence of iconic objects which play crucial roles in scene classification}.

  In CNN, the convolutional features are pooled and then transformed nonlinearly layer by layer before feeding to the FC layer. Low-level convolutional layers perform like Gabor filters and color blob detectors \cite{yosinski2014transferable}, and mainly capture the \emph{edges} and/or \emph{textures} information. During the forward layer-wise process of the CNN, the features exhibit more abstractive meaning, and become more robust to local image variations. The FC layers significantly reduce the dimension of the convolutional features, avoiding huge memory and computation cost.
  On the other hand, the high-level nature of the FC-features makes them difficult to extract strong local subtle structures of the images, such as fine-scale objects or their parts. This fact can be also verified in recent work \cite{mahendran2014understanding}, where the authors shown that the images reconstructed from the FC-features can preserve global layouts of the original images, but they are very fuzzy, losing fine-grained local details and even the positions of the parts. By contrast, the reconstructions from the convolutional features are much more photographically faithful to the original ones. Therefore, the FC-features may not well capture the local object information and fine structures, while these mid-level features are of great importance for scene classification. To  illustrate the complementary capabilities of the two features, we show the classification results by each of them in Fig \ref{fig:ConvFC}. It can be found that the two types of features are capable of discriminating different scene categories by capturing either local subtle objects information or global structures of the images, providing strong evidence that the convolutional features are indeed beneficial.


   To further illustrate the challenge of scene classification, we present several pairs of ambiguous scene categories (from the MIT Indoor 67) in Fig. \ref{fig3}. The images in each category pair exhibit relatively similar global structure and layout, but have main difference in representative local objects or specific regions. For each pair, we train a SVM classifier with the FC-features, the convolutional features extracted from the $4^{th}$ layer, or their combination. The classification errors on the test sets are summarized in bottom table in Fig. \ref{fig3}. As can be observed, the FC-features do not perform well on these ambiguous category pairs, while the convolutional features yield better results by capturing more local differences. As expected, combination of them eventually leads to performance boost by computing both global and local image structures. It achieves zero errors on three category pairs which have strong local discriminants between them, e.g. $jewellery$ vs $shoe$.

   To further investigate the different properties of the FC-features and convolutional features, we calculate the statistics of their activations on the MIT Indoor 67. We record the top 1,000 images which have the largest average activations in the last FC layer and the $4^{th}$ convolutional layer, respectively. Fig. \ref{fig7} shows the distributions of these 1,000 images among 67 categories. As can be seen, there exist obvious difference between two distributions, implying that the representation abilities of the two features are varied significantly across different scene categories. It also means that some scene categories may include strong characteristics of the FC-features, while the others may be more discriminative with the convolutional features.
   These results, together with previous discussions, can readily lead to a conclusion that \textit{the FC-features and convolutional features can be strongly complementary to each other, and  both global layout and local fine structure are crucial to yield a robust yet discriminative scene representation}.
   \begin{figure*}
\centering
\includegraphics[scale=0.5]{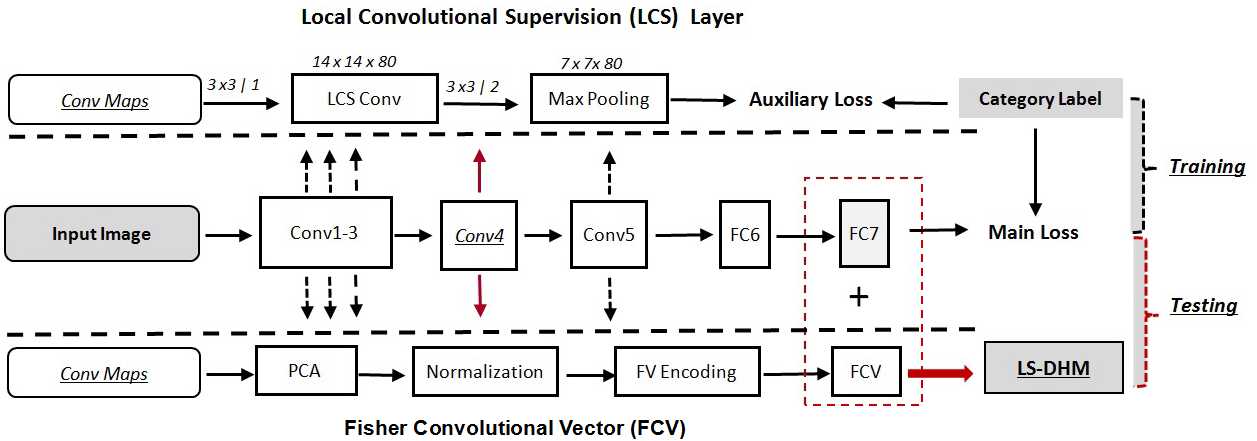}
\caption{The structure of Locally-Supervised Deep Hybrid Model (LS-DHM) built on 7-layer AlexNet \cite{krizhevsky2012imagenet}. The LS-DHM can be constructed by incorporating the FCV with external FC-features from various CNN models, such as GoogleNet \cite{szegedy2014going} or VGGNet \cite{simonyan2014very}.}
\label{fig10}
\end{figure*}
%
%

\subsection{Locally-Supervised Deep Hybrid Model}
 In this subsection, we present the details of the proposed Locally-Supervised Deep Hybrid Model (LS-DHM), which incorporates both the FCV representation and FC-features of the CNN. The structure of the LS-DHM is presented in Fig. \ref{fig10}. It is built on a classical CNN architecture, such as the AlexNet \cite{krizhevsky2012imagenet}  or the Clarifai CNN \cite{zeiler2014visualizing}, which has five convolutional layers followed by another two FC layers.

 \textbf{Local Convolutional Supervision (LCS)}. We propose the LCS to enhance the local objects and fine structures information in the convolutional layers. Each LCS layer is directly connected to one of the convolutional layers in the main CNN. Specifically, our model can be formulated as follows. Given $N$ training examples, $\{\textbf{I}_i,y_i\}_{i=1}^N$, where $\textbf{I}_i$ demotes a training image, and $y_i$ is the label, indicating the category of the image.  The goal of the conventional CNN is to minimize,
\begin{equation} \label{Eq:TraditionalCNN}
   \arg\min_{\textbf{W}}{\sum_{i=1}^{N}{\mathcal{L}(y_i, f(\textbf{I}_i;\textbf{W})) + \|\textbf{W}\|_2}}
\end{equation}
where $\textbf{W}$ is model weights that parameterize the function $f(\textbf{x}_i;\textbf{W})$ . $\mathcal{L}(\cdot)$ denotes the loss function, which is typically a hinge loss for our classification task. $\|\textbf{W}\|_2$ is the regularization term. The training of the CNN is to look for a optimized $\textbf{W}$ that maps $\emph{I}_i$ from the image space onto its label space.

Extending from the standard CNN, the LCS introduces a new auxiliary loss ($\ell^a$) to the convolutional layer of the main networks, as shown in Fig. \ref{fig10}. It can be formulated as,
\begin{equation}\label{MTL_general}
 \arg\min_{\textbf{W},\textbf{W}^a}\!\!{\sum_{i=1}^{N}\!{\mathcal{L}(y_{i}\!,\!f(\textbf{I}_i\!;\!\textbf{W})) } \!\!+\!\! \sum_{i=1}^{N}\!\!{\sum_{a\in A}\!\!{\lambda^a\ell^a(y^a\!, \!f(\textbf{I}_i;\!\textbf{W}^a ))} }},
\end{equation}
where $\ell^a$ is auxiliary loss function, which has the same form as the main loss $\mathcal{L}$ by using the hinge loss. $\lambda^a$ and $\textbf{W}^a$ denote the importance factor and model parameters of the auxiliary loss. Here we drop the regularization term for notational simplicity. Multiple auxiliary loss functions can be applied to a number of convolutional layers selected in set $A$, allowing our design to build multiple LCS layers upon different convolutional layers.  In our model, $\textbf{W}$ and $\textbf{W}^a$ share the same parameters in the low convocational layers of the main CNN, but have independent parameters in the high-level convolutional layers or the FC layers. The label used for computing the auxiliary loss is the same as that of the main loss, $y^a_i=y_i$, allowing the LCS to propagate the final label information to the convolutional layers in a more direct way. This is different from recent work on exploring the CNN model for multi-task learning (MTL) (e.g. for face alignment \cite{Zhang2015} or scene text detection \cite{He2015} ), where the authors applied completely different supervision information to various auxiliary tasks in an effort to facilitate the convergence of the main task.

By following the conventional CNN, our model is trained with the classical SGD algorithm w.r.t $\textbf{W}$ and $\textbf{W}^a$. The structure of our model is presented in Fig. \ref{fig10}, where the proposed LCS is built on just one convolutional layer (the $4^{th}$ layer) of the main CNN. Similar configuration can be readily extended to multiple convolutional layers. The LCS contains a single convolutional layer followed by a max pooling operation. We apply a small-size kernel of $3\times3$ with the stride of 1 for the convolutional layer, which allows it to preserve the local detailed information as much as possible. The size of the pooling kernel is set to $3 \times 3$, with the stride of 2. The feature maps generated by the new convolutional and pooling layers have the sizes of $14\times14\times80$ and  $7\times7\times80$ respectively, compared to the $14\times14\times384$ feature maps generated by the $4^{th}$ layer of the main CNN.

In particular, the pooling layer in the LCS is directly connected to the final label in our design, without using any FC-layer in the middle of them.  This specific design encourages the activations in the convolutional layer of the LCS to be directly predictive of the final label. Since each independent activation in convolutional layer may include meaningful local semantics information (e.g. local objects or textures located within its receptive field),  further correlating or compressing these activations through a FC layer may undermine these fine-scale but local discriminative information. Thus our design provides a more principled approach to recuse these important local cues by enforcing them to be directly sensitive to the category label. This design also sets the LCS apart from related convolutional supervision approaches developed in \cite{szegedy2014going,Yang2015,wang2015training,lee2014deeply}, where the FC layer
is retained in the auxiliary supervision layers. Furthermore, these related approaches only employ the FC-features for image representation, while our method explores both the convolutional features and the FC-features by further developing an efficient FCV descriptor for encoding the convolutional features.



  \textbf{Fisher Convolutional Vector (FCV)}. Although the local object and region information in the convolutional layers can be enhanced by the proposed LCS layers, it is still difficult to preserve  these information sufficiently  in the FC-representation, due to multiple hierarchical compressions and abstractions. A straightforward approach is to directly employ all these convolutional features for image description.  However, it is non-trivial to directly apply them for training a classifier. The convolutional features are computed densely from the original image, so that they often have a large number of feature dimensions, which may be significantly redundant. Furthermore, the densely computing also allows the features to preserve explicit spatial information of the image, which is not robust to various geometric deformations.

  Our goal is to develop a discriminative mid-level representation that robustly encodes rich local semantic information in the convolutional layers.
  Since each activation vector in the convolutional feature maps has a corresponding receptive field (RF) in the original image, this allows it to capture the local semantics features within its RF, e.g. fine-scale objects or regions. Thus the activation vector can be considered as an independent mid-level representation regardless of its global spatial correlations. For the scene images, such local semantics are of importance for fine-grained categorization, but are required to increase their robustness by discarding explicit spatial information. For example, the images of the \emph{car} category may include various numbers and multi-scale cars in complectly different locations.
  Therefore, to improve the robustness of the convolutional features without degrading their discriminative power, we develop the FCV representation that computes the orderless mid-level features by leveraging the Fisher Vector (FV) encoding \cite{Jaakkola1998,sanchez2013image}.

  The Fisher Kernel \cite{Jaakkola1998} has been proven to be extremely powerful for pooling a set of dense local features (e.g. SIFT \cite{lowe2004distinctive}), by removing global spatial information \cite{sanchez2013image}. The convolutional feature maps can be considered as a set of dense local features, where each activation vector works as a feature descriptor. Specifically, given a set of convolutional maps with the size of $H \times W \times D$ (from a single CNN layer), where $D$ is the number of the maps (channels) with the size of $H \times W$, we get a set of $D$-dimensional local convolutional features ($\textbf{C}$),
\begin{eqnarray}
  \textbf{C}=\{C_1,C_2,...,C_T\}, T=H\times W
\end{eqnarray}
where $\textbf{C}\in\mathbb{R}^{D\times T}$. $T$ is the number of local features which are spatially arranged in $ H\times W$. To ensure that each feature vector contributes equally and avoid activation abnormity, we normalize each feature vector into interval [-1, 1] by dividing its maximum magnitude value  \cite{Wang2015},
\begin{eqnarray}
  C_t=C_t/\max\{|C_t^1|,|C_t^2|,...,|C_t^D|\}
\end{eqnarray}

\renewcommand{\algorithmicrequire}{ \textbf{Input:}} 
\renewcommand{\algorithmicensure}{ \textbf{Output:}} 

\begin{algorithm}[tb]
\caption{\small Compute FCV from the Convolutional Maps}
\label{alg:alg1}
\begin{algorithmic}[1] 
\REQUIRE ~~\\ 
Convolutional features maps with the size of $H \times W \times D$. \\
GMM parameters, $\lambda =\{\omega_{k},\mu_{k},\sigma_{k},k=1,\ldots,K \}$.
 \\
\ENSURE ~~\\ 
FCV with $2MK$ dimensions.\\

\underline{\textbf{Step One:}} Extract Local Convolutional Features.
\STATE Get $T=H\times W$ normalized feature vectors, $\textbf{C}\in\mathbb{R}^{D\times T}$.
\STATE Reduce dimensions using PCA,  $\hat{\textbf{C}}\in\mathbb{R}^{M\times T}$, $M<D$.\\

\underline{\textbf{Step Two:}} Compute the FV Encoding.

\STATE Compute the soft assignment of $\hat{C}_t$ to Gaussian $k$:\\
$\gamma_t^k=\frac{\omega_{k}\mu_{k}(\hat{C}_t)}{\sum_{j=1}^K\omega_{j}\mu_{j}(\hat{C}_t)}$, $k=1,\ldots,K$.\\
\STATE Compute Gaussian accumulators:\\
$S_{k}^0=\sum_{t=1}^T\gamma_t^k$, $S_{k}^{\mu}=\sum_{t=1}^T\gamma_t^k\hat{C}_t$, $S_{k}^{\sigma}=\sum_{t=1}^T\gamma_t^k\hat{C}_t^2$.\\
where $S_{k}^0\in\mathbb{R}$, and $S_{k}^\mu,S_{k}^{\sigma}\in\mathbb{R}^M$, $k=1,\ldots,K$.\\

\STATE Compute FV gradient vectors:\\
$F_k^{\mu}=(S_k^\mu-\mu_kS_k^0)/(\sqrt{\omega_{k}\sigma_{k}})$\\
$F_k^{\sigma}=(S_k^\sigma-2\mu_kS_k^\mu+(\mu_k^2-\sigma_k^2)S_k^0)/(\sqrt{2\omega_{k}\sigma_{k}^2})$\\
where $F_k^\mu,F_k^\sigma \in \mathbb{R}^M$, $k=1,\ldots,K$.\\
\STATE Concatenate two gradient vectors from $K$ mixtures:\\
$FCV=\{F_1^{\mu},...,F_K^{\mu},F_1^{\sigma},...,F_K^{\sigma}\} \in \mathbb{R}^{2MK}$. \\
\STATE Implement power and $\ell_2$ normalization on the FCV.
\end{algorithmic}
\end{algorithm}

We aim to pool these normalized feature vectors to achieve an image-level representation.  We adopt the Fisher Vector (FV) encoding \cite{sanchez2013image} which  models the distribution of the features by using a Gaussian Mixture Model (GMM), and  describe an image by considering the gradient of likelihood w.r.t the GMM parameters, i.e. mean and covariance. By following previous work \cite{sanchez2013image}, we first apply the Principal Component Analysis (PCA) \cite{Jolliffe2002} for reducing the number of feature dimensions to $M$. For the FV encoding, we adopt a GMM with $K$ mixtures, $G_{\lambda}=\{g_{k},k=1\ldots K\}$, where $\lambda =\{\omega_{k},\mu_{k},\sigma_{k},k=1\ldots K \}$. For each GMM mixture, we compute two gradient vectors, $F_k^{\mu}\in \mathbb{R}^M$ and $F_k^{\sigma}\in \mathbb{R}^M$, with respect to the means and standard deviations respectively. The final FCV representation is constructed by concatenating two gradient vectors from all mixtures, which results in an orderless $2MK$-dimensional representation. The FCV can be feed to a standard classifier like SVM for classification. Note that the dimension number of the FCV is fixed, and is independent to the size of the convolutional maps, allowing it to be directly applicable to various convolutional layers. Details of computing the FCV descriptor is described in Algorithm 1.

  \textbf{Locally-Supervised Deep Hybrid Model (LS-DHM)}. As discussed, scene categories are defined by multi-level image contents, including the mid-level local \emph{textures} and \emph{objects}, and the high-level \emph{scenes}. While these features are captured by various layers of the CNN, it is natural to integrate the mid-level FCV (with LCS enhancement) with the high-level FC-features by simply concatenating them, which forms our final LS-DHM representation. This allows scene categories to be coarsely classified by the FC-features with global structures, and at the same time, many ambiguous categories can be further discriminated finely by the FCV descriptor using local discriminative features. Therefore, both types of features compensate to each other, which leads to performance boost.

  The structure of the LS-DHM is shown in Fig. \ref{fig10}. Ideally, the proposed FCV and LCS are applicable to multiple convolutional layers or deeper CNN models.  In practice, we only use the single convolutional layer (the $4^{th}$ layer) in the celebrated 7-layer AlexNet for computing the FCV in current work. This makes the computation of FCV very attractive, by only taking about $60ms$ $per$ $image$ on the SUN379 by using a single GPU. Even that we has achieved very promising results in the current case, and better performance can be expected by combining the FCV from multiple layers, which will be investigated in our future work. Furthermore, the construction of the LS-DHM is flexible  by integrating the FCV with various FC-features of different CNNs, such as the AlexNet \cite{krizhevsky2012imagenet}, GoogleNet \cite{szegedy2014going} and VGGNet \cite{simonyan2014very}. The performance of the LS-DHM are varied by various capabilities of FC-features.

  The LS-DHM representation is related to the MOP-CNN \cite{gong2014multi}, which extracts the local features by computing multiple FC-features from various manually-divided local image patches. Each FC-feature of the MOP-CNN is analogous to an activation vector in our convolutional maps. The FCV captures richer local information by densely scanning the whole image with the receptive fields of the activation vectors, and providing a more efficient pooling scheme that effectively trades off  the robustness and discriminative ability. These advantages eventually lead to considerable performance improvements over the MOP-CNN. For example, our LS-DHM achieved 58.72\% (\emph{vs} 51.98\% by MOP-CNN) on the SUN397 and 73.22\% (\emph{vs} 68.88\% by MOP-CNN) on the MIT Indoor76, by building on the same AlexNet architecture.
  Furthermore, the FCV and FC-features of the LS-DHM share the same CNN model,  making it significantly more efficient by avoiding repeatedly computing the network, while the MOP-CNN repeatedly implements the same network 21 times to compute all 3-level local patches \cite{gong2014multi}.
  In addition, the LS-DHM representation is flexible to integrate the FCV with more powerful FC-features, leading to further performance improvements, as shown in Section IV.


\section{Experimental Results and Discussions}
The performance of the proposed LS-DHM is evaluated on two heavily benchmarked scene datasets: the MIT Indoor67 \cite{quattoni2009recognizing} and the SUN397 \cite{xiao2010sun}. We achieve the best performance ever reported on both benchmarks.

 \textbf{The MIT Indoor67 \cite{quattoni2009recognizing}} contains 67 indoor-scene categories and a total of 15,620 images, with at least 100 images per category. Following the standard evaluation protocol of \cite{quattoni2009recognizing}, we use 80 images from each category for training, and another 20 images for testing.  Generally, the indoor scenes have strong object information, so that they can be better discriminated by the iconic objects they contain, such as the $bed$ in the $bedroom$ and the $table$ in the $dinning room$.

 \textbf{The SUN397 \cite{xiao2010sun}} has a large number of scene categories by including 397 categories and totally 108,754 images. This makes it extremely challenging for this task. Each category has at least 100 images.  We follow the standard evaluation protocol provided by the original authors \cite{xiao2010sun}. We train and test the LS-DHM on ten different partitions, each of which has
50 training and 50 test images.  The partitions are fixed and publicly available from \cite{xiao2010sun}. Finally the average classification accuracy of ten different tests is reported.
\subsection{Implementation Details }

We discuss the parameters of FCV descriptor, and various CNN models which are applied for computing the FC-features of our LS-DHM. For the FCV parameters, we investigate the number of reduced dimensions by PCA, and the number of Gaussian mixtures for FV encoding.
The FCV is computed from the $4^{th}$ convolutional layer with the LCS enhancement, building on the 7-layer AlexNet architecture. The performance of the FCV computed on various convolutional layers will be evaluated below. The LS-DHM can use various FC-features of different CNN models, such as the AlexNet \cite{krizhevsky2012imagenet}, GoogleNet \cite{szegedy2014going} and VGGNet \cite{simonyan2014very}. We refer the LS-DHM with different FC-features as LS-DHM (AlexNet), LS-DHM (GoogleNet) and LS-DHM (VGGNet). All deep  CNN models in our experiments are trained with the large-scale Places dataset \cite{zhou2014learning}. Following previous work \cite{gong2014multi,zhou2014learning}, the computed LS-DHM descriptor is feeded to a pre-trained linear SVM for final classification.


\begin{figure}
\centering
\subfigure [PCA Dimension Reductions]{\includegraphics[scale=0.32]{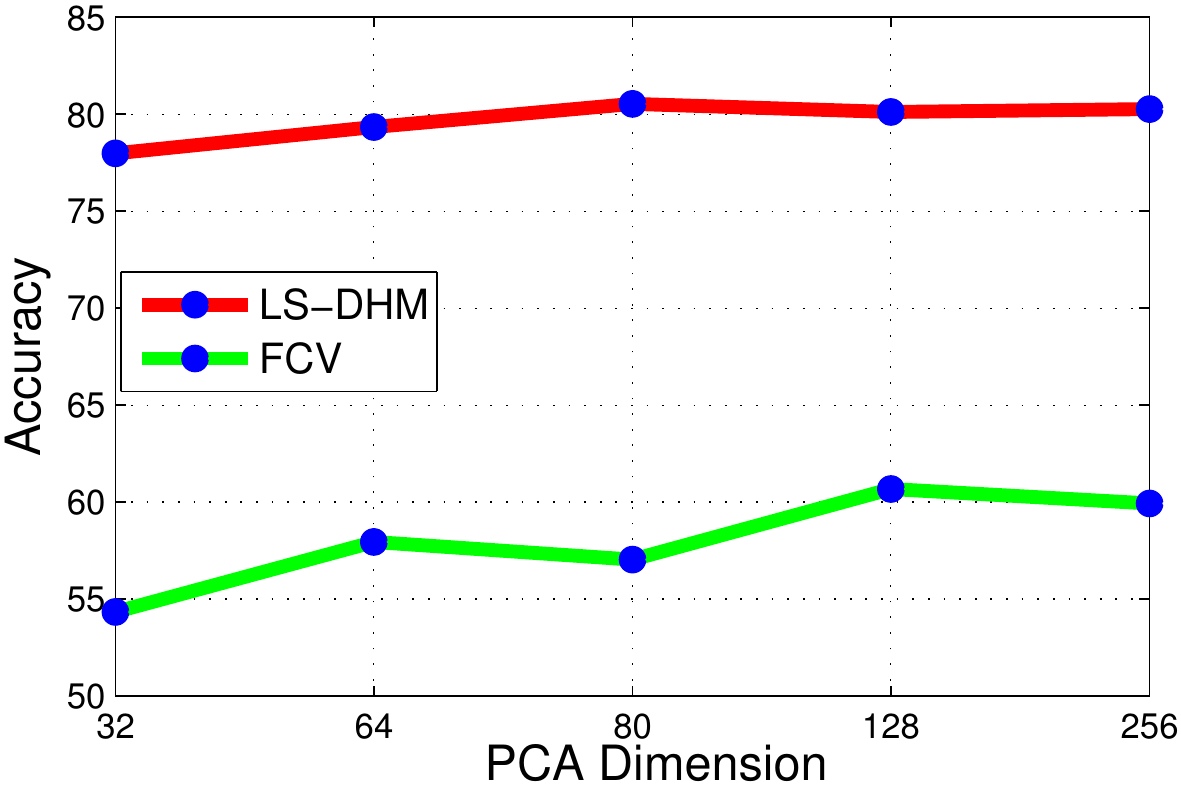}}
\subfigure[Gaussian Mixtures]{\includegraphics[scale=0.32]{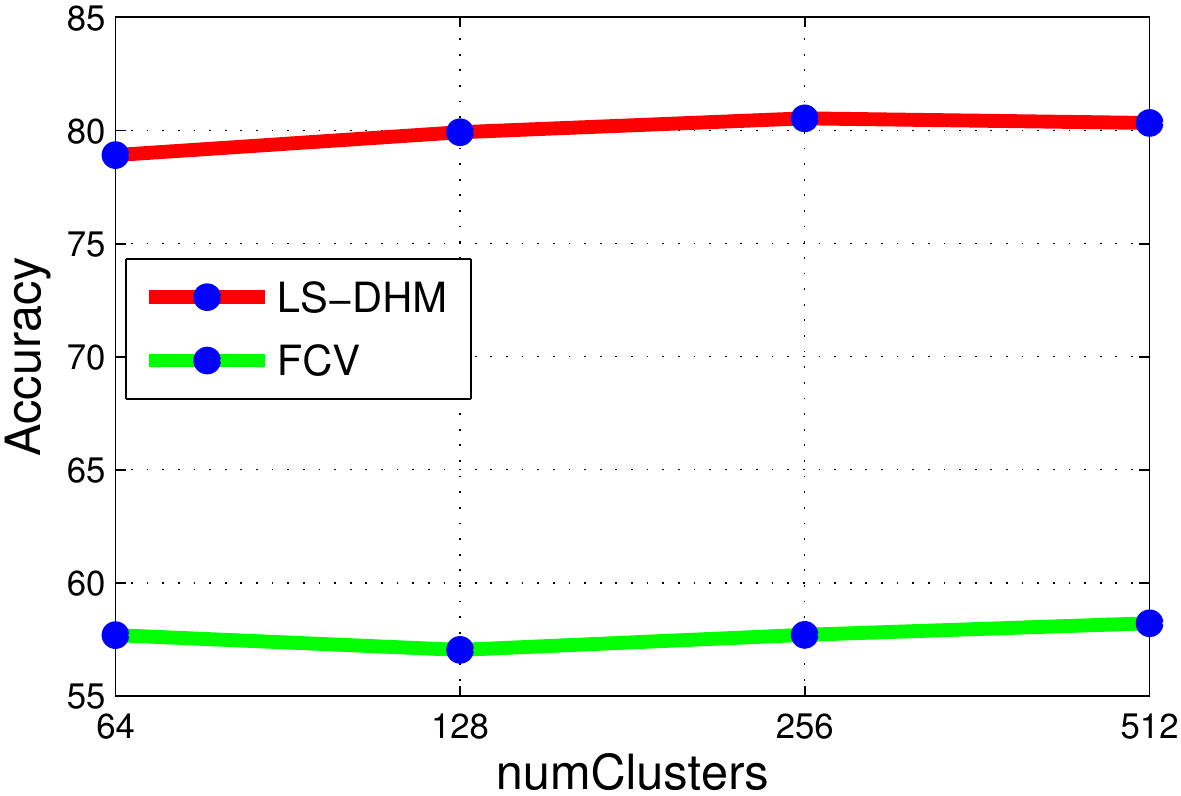}}

\caption{The performance of the FCV and LS-DHM (GoogleNet) with various numbers of  (left) reduced dimensions, and (right) the Gaussian mixtures. Experiments were conducted on the MIT Indoor67.}
\label{fig4}
\end{figure}

\begin{figure}
\centering
\includegraphics[scale=0.4]{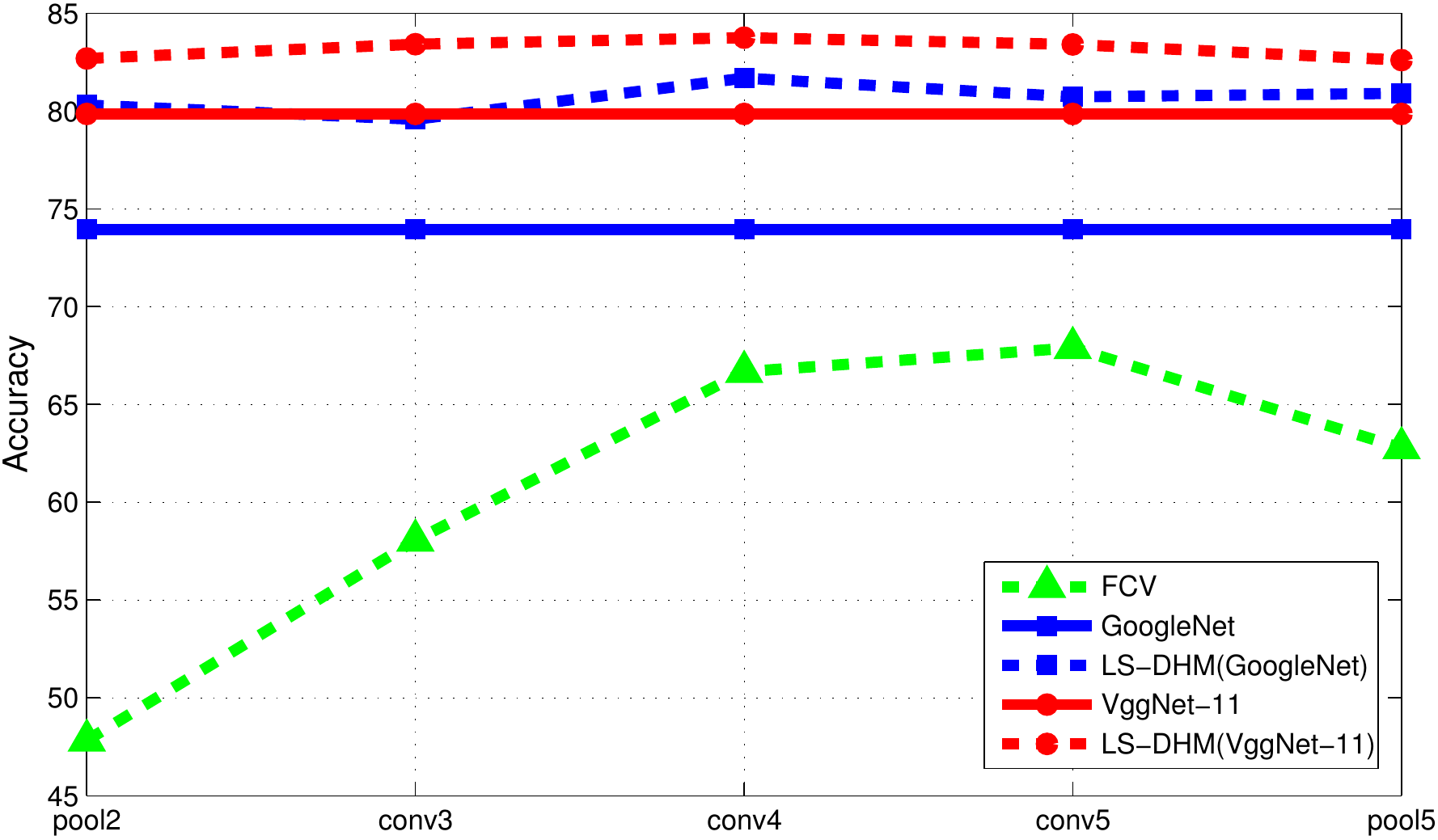}

\caption{Performance of the FCV computed at various convolutional layers of the AlexNet, and the LS-DHM with different FC-features from the GoogleNet or VGGNet. The experiments were conducted on the MIT Indoor67.
}
\label{fig5}
\end{figure}


    \textbf{Dimension reduction}. The $4^{th}$ convolutional layer of the AlexNet includes 384 feature maps,
    which are transformed to a set of 384D convolutional features. We verify the effect of the dimension reduction (by using PCA) on the performance of the FCV and LS-DHM. The numbers of retained dimensions are varied from 32 to 256, and experimental results on the MIT Indoor67 are presented in the left of Fig. \ref{fig4}. As can be found, the number of retained dimensions does not impact the performance of FCV or LS-DHM significantly. By balancing the performance and computational cost, we choose to retain 80 dimensions for computing the FCV descriptor in all our following experiments.

    \textbf{Gaussian mixtures}. The FV encoding requires learning the GMM as its dictionary. The number of the GMM mixtures also impact the performance and the complexity of FCV. Generally speaking, larger number of the Gaussian mixtures leads to a stronger discriminative power of the FCV, but at the cost of using more FCV dimensions. We investigate the impact of the mixture number on the FCV and LS-DHM by varying it from 64 to 512. We report the classification accuracy on the MIT Indoor67 in the right of Fig. \ref{fig4}. We found that the results of FCV or LS-DHM are not very sensitive to the  number of the mixtures, and finally used 256 Gaussian mixtures for our FCV.


\begin{figure*}
\centering
\subfigure{\includegraphics[height=2.2in,width=7in,angle=0]{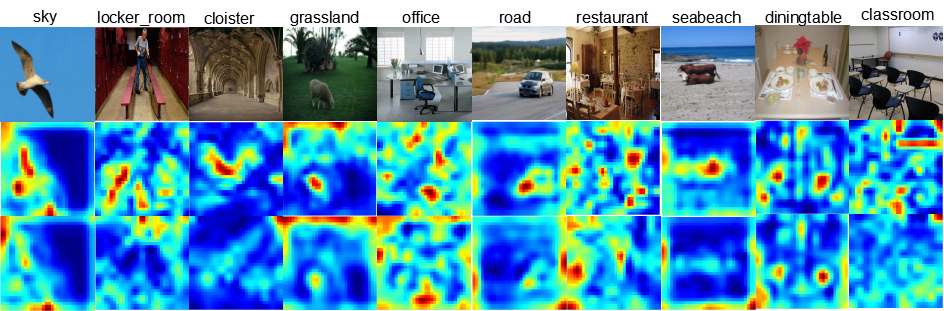}}

\caption{Comparisons of the convolutional maps (the mean map of 4th-convolutional layer) with the LCS enhancement (middle row), and without it (bottom two). The category name is list on the top of each image. Obviously, the LCS enhances the local object information in the convolutional maps significantly.  These object information are crucial to identify those scene categories, which are partly defined by some key objects.}
\label{fig:featuremap_lcs}
\end{figure*}

\begin{table}[tb]
\centering
\caption{Comparisons of various pooling methods on the MIT Indoor67. The LS-DHM is constructed by integrating the FC-features of GoogleNet and the encoded convolutional features, computed from  AlexNet  with or without (w/o)  LCS layer.}
\begin{tabular}{l||c|c||c|c||c|c}
\hline
Encoding & \multicolumn{2}{c||}{Conv-Features Only}& \multicolumn{2}{c||}{FC-Features} &\multicolumn{2}{c}{LS-DHM}\\
\cline{2-3}
Method & w/o LCS & LCS &\multicolumn{2}{c||}{GoogleNet}& w/o LCS &  LCS\\
 \hline\hline
 Direct & 51.46 & 58.41 & \multicolumn{2}{c||}{}& 76.95&77.40\\
 BoW &37.28 & 57.38 &\multicolumn{2}{c||}{73.79}& 78.09& 78.64\\
 FCV& \textbf{57.04} & \textbf{65.67}& \multicolumn{2}{c||}{}&\textbf{80.34}&\textbf{81.68}\\ \hline

\end{tabular}

\label{encoding}
\end{table}

 \subsection{Evaluations on the LCS, FCV and LS-DHM}

  We investigate the impact of individual LCS  or FCV to the final performance.
  The FC-features from the GoogleNet or VGGNet are explored to construct the LS-DHM representation.

\textbf{On various convolutional layers}. The FCV can be computed from various convolutional layers, which capture the feature abstractions from low-level to mid-level, such as \emph{edges}, \emph{textures} and \emph{objects}. In this evaluation, we investigate the performance of FCV and the LS-DHM on different convolutional layers, with the LCS enhancement. The results on the AlexNet, from the \emph{Pool2} to \emph{Pool5} layers, are presented in Fig. \ref{fig5}. Obviously, both FCV and LS-DHM got the best performance on the $4^{th}$ convolutional layer. Thus we select this layer for building the LCS layer and computing the FCV. By integrating the FCV, the LS-DHMs achieve remarkable performance improvements over the original VGGNet or GoogleNet, demonstrating the efficiency of the proposed FCV.  Besides, we also investigate performance of the FCV by computing it from multiple convolutional layers. The best performance is achieved at 83.86\%, by computing the FCV from \textit{conv4}, \textit{conv5} and \textit{pool5}. However, this marginal improvement results in three times of feature dimensions, compared to the FCV computed from single \textit{conv4}. Therefore, by trading off the performance and computational cost,  we use single \textit{conv4} to compute our FCV in all following experiments. Notice that using more convolutional layers for the FCV dose not improve the performance further, i.e., computing the FCV from \textit{conv3-5} and \textit{pool5} results in a slight reduction in performance, with 83.41\%.



\textbf{On the pooling approaches}. We further evaluate the FCV by investigating various pooling approaches for encoding the convolutional features.
 We compare the FV encoding with direct concatenation method and the BoW pooling \cite{Sivic2003,Csurka2004}. The results on the MIT Indoor67 are shown in Table \ref{encoding}. As can be seen, the FCV achieves remarkable improvements over the other two approaches, especially on purely exploring the convolutional features where rough global structure is particularly important. In particular, the BoW without the LCS yields a low accuracy of $37.28\%$.
 It may due to the orderless nature of BoW pooling which completely discarding the global spatial information.
 The convolutional features trained without the LCS are encouraged to be abstracted to the high-level FC features. This enforces the convolutional features to be globally-abstractive by preserving rough spatial information for high-level scene representation.
 On the contrary, the direct concatenation method preserves explicit spatial arrangements, so as to obtain a much higher accuracy.  But the explicit spatial order is not robust to local distortions, and it also uses a large amount of feature dimensions. The FV pooling increases the robustness by relaxing the explicit spatial arrangements; and at the same time, it explores more feature dimensions to retain its discriminative power,  leading to a performance improvement.

\textbf{On the LCS}. As shown in Table \ref{encoding}, the LCS improves the performance of all pooling methods substantially by enhancing the mid-level local semantics (e.g. \emph{objects} and \emph{textures}) in the convolutional layers. The accuracy by the BoW is surprisingly increased to $57.38\%$ with our LCS enhancement. The performance is comparable to that of the direct concatenation which uses a significant larger number of feature dimensions. One of the possible reasons may be that the LCS enhances the local object information by directly enforcing the supervision on each activation in the convolutional layers, allowing the image content within RF of the activation to be directly predictive to the category label. This encourages the convolutional activations to be locally-abstractive, rather than the globally-abstractive in conventional CNN.
These locally-abstractive convolutional features can be robustly identified without their spatial arrangements, allowing them to be discriminated by the orderless BoW representation. As shown in Fig. \ref{fig:featuremap_lcs}, our LCS significantly enhances the local object information in the convolutional maps, providing important cues to identify those categories, where some key objects provide important cues. For example, strong \emph{head} information is reliable to recognize the \emph{person} category, and confident \emph{plate} detection is important to identify a \emph{diningtable} image.

\begin{figure*}
\centering
\subfigure{\includegraphics[scale=0.55]{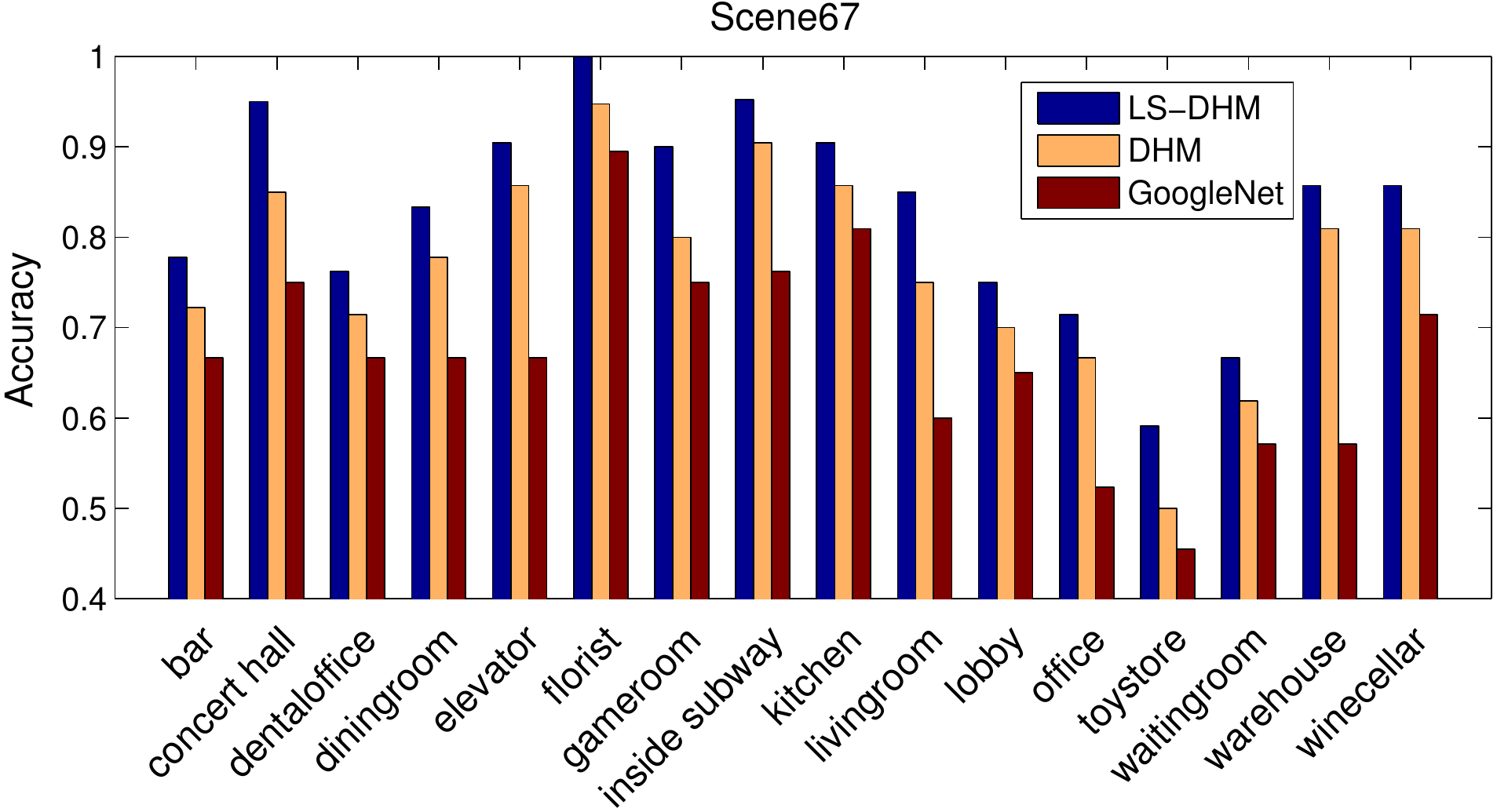}}
 \subfigure{\includegraphics[scale=0.5]{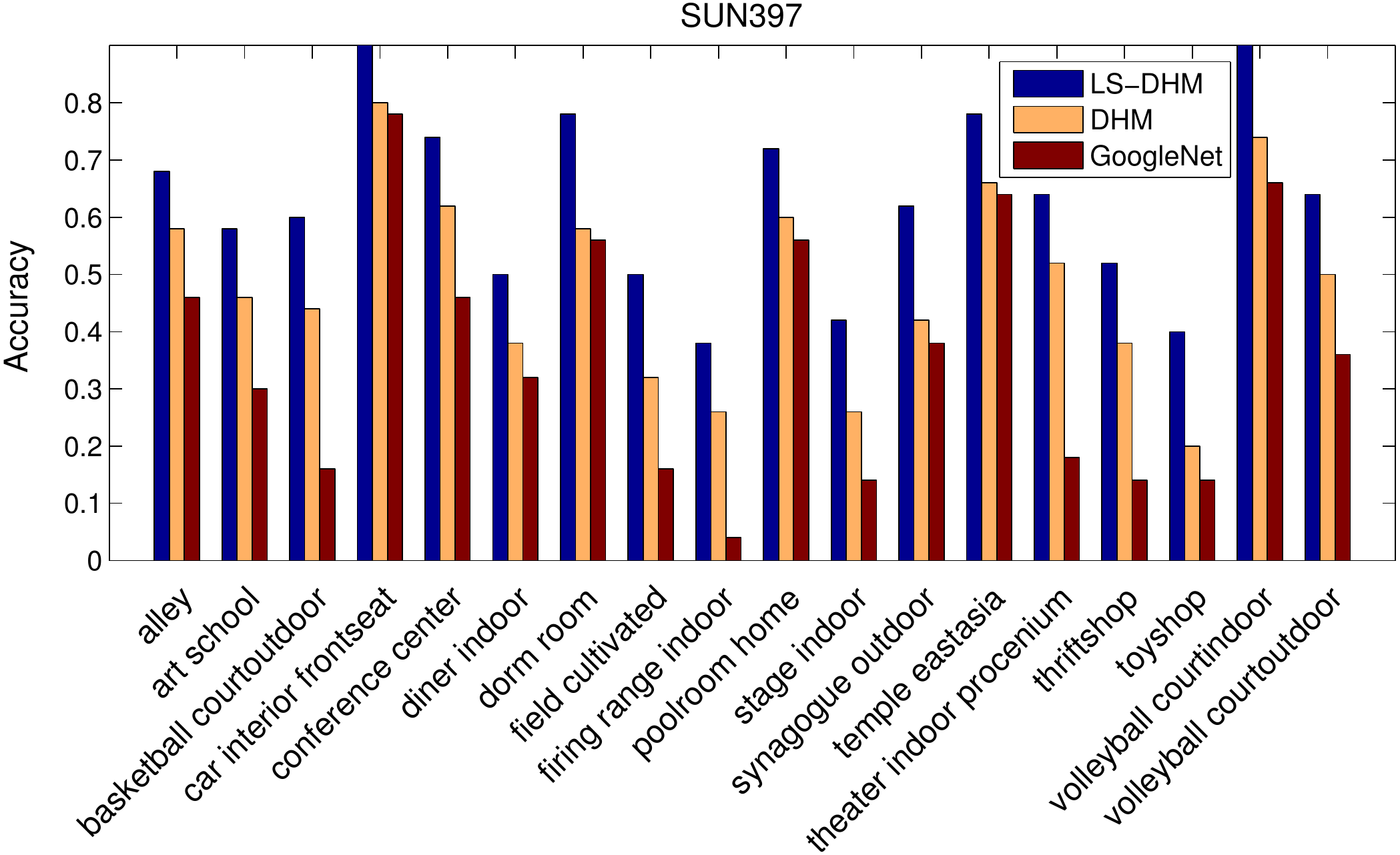}}

\caption{Classification accuracies of several example categories with FC-features (GoogleNet), DHM and LS-DHM on the MIT Indoor67 and SUN397. DHM denotes the LS-DHM without LCS enhancement.}
\label{fig6}
\end{figure*}

\textbf{On the LS-DHM}. In the Table \ref{encoding}, the single FC-features yield better results than the convolutional features, suggesting that scene categories are primarily discriminated by the global layout information.
Despite capturing rich fine-scale semantics, the FCV descriptor  perseveres little global spatial  information by using the FCV pooling. This reduces its discriminative ability to identify many high-level (e.g. \emph{scene}-level) images, so as to harm its performance.
However, we observed that, by intergrading both types of features, the proposed LS-DHM archives remarkable improvements over the individual FC-features in all cases. The largest gain achieved by our LS-DHM with the LCS improves the accuracy of individual FC-features from 73.79\% to 81.68\%. We got a similar large improvement on the SUN397, where our LS-DHM develops the strong baseline of GoogleNet considerably, from 58.79\% to 65.40\%.
Furthermore, these facts are depicted more directly in Fig. \ref{fig6}, where we show the classification accuracies of various features on a number of scene categories from the MIT Indoor67 and SUN397. The significant impacts of the FCV and LCS to performance improvements are shown clearly.
These considerable improvements convincingly demonstrate the strong complementary properties of the convolutional features and the FC-features, giving strong evidence that the proposed FCV with LCS is indeed beneficial to scene classification.

\textbf{On computational time}. In test processing, the running time of LS-DHM includes computations of the FC-feature (CNN forward propagation) and FCV, which are about \textit{61ms} (by using a single TITAN X GPU with the VGGNet-11) and \textit{62ms} (CPU time) per image, respectively.  The time of FCV can be reduced considerably by using GPU parallel  computing. The LCS is just implemented in training processing, so that it dose not raise  additional computation in the test. For training time, the original VGGNet-11 takes about 243 hours (with 700,000 iterations) on the training set of Place205, which is increased slightly to about 262 hours by adding the LCS layer (on the \textit{conv4}). The models were trained by using 4 NVIDIA TITAN X GPUs.

\subsection{Comparisons with the state-of-the-art results}


We compare the performance of our LS-DHM with recent approaches on the MIT Indoor67 and SUN397. The FCV is computed from the AlexNet with LCS. 
Our LS-DHM representation is constructed by integrating the FCV with various FC-features of different CNN models. The results are compared extensively in Table \ref{table2} and \ref{table3}.

 The results show that our LS-DHM with the FC-features of 11-layer VGGNet outperforms all previous Deep Learning (DL) and FV methods substantially on both datasets.  For the DL methods, the Places-CNN trained on the Place data by Zhou \emph{et al.} \cite{zhou2014learning} provides strong baselines for this task. Our LS-DHM, building on the same AlexNet, improves the performance of Places-CNN with a large margin by exploring the enhanced convolutional features. It achieves about 10\% and 8\% improvements over the Places-CNN on the MIT Indoor67 and SUN397 respectively. These considerable improvements confirm the significant impact of FCV representation which captures important mid-level local semantics features for discriminating many ambiguous scenes.

\begin{table}[tb]
\centering  
\caption{Comparisons of the proposed LS-DHM with the state-of-the-art on the MIT Indoor67 database.}
\begin{tabular}{p{3.3cm}||p{1.6cm}<{\centering}||p{1.6cm}<{\centering}}
\hline
\textbf{Method} & \textbf{Publication} & \textbf{Accuracy}($\%$)\\
\hline
Patches+Gist+SP+DPM\cite{singh2012unsupervised}& ECCV2012 &49.40\\
BFO+HOG\cite{kobayashi2013bfo} & CVPR2013&58.91\\
FV+BoP\cite{juneja2013blocks}&CVPR2013 &63.10\\
FV+PC\cite{doersch2013mid} &  NIPS2013 &68.87\\
FV(SPM+OPM)\cite{xie2014orientational}& CVPR2014 &63.48\\
DSFL\cite{zuo2014learning} &ECCV2014& 52.24\\
LCCD+SIFT \cite{Guo2015} &arXiv2015&65.96\\

 \hline
DSFL+CNN\cite{zuo2014learning} &ECCV2014& 76.23\\

CNNaug-SVM\cite{razavian2014cnn}& CVPR2014&69.00\\
MOP-CNN \cite{gong2014multi} &ECCV2014 &68.90\\
MPP \cite{Yoo2015}& CVPR2015&77.56\\
MPP \cite{Yoo2015}+DSFL\cite{zuo2014learning}&CVPR2015&80.78\\
FV-CNN (VGGNet19)\cite{cimpoi2015deep}&CVPR2015 &81.00\\
DAG-VGGNet19 \cite{Yang2015}&ICCV2015&77.50\\
C-HLSTM \cite{Zuo2015} &arXiv2015&75.67\\
Ms-DSP (VGGNet16) \cite{Gao2015} &arXiv2015&78.28\\
\hline\hline

Places-CNN(AlexNet)\cite{zhou2014learning} &NIPS2014& 68.24\\
LS-DHM(AlexNet) &--& 78.63\\
\hline
 GoogleNet &--& 73.96\\
LS-DHM(GoogleNet)&--& 81.68\\
 \hline
VGGNet11 &--& 79.85\\
LS-DHM(VGGNet11)&--& \textbf{83.75}\\

\hline
\end{tabular}

\label{table2}
\end{table}

\begin{table}[tb]
\centering  
\caption{Comparisons of the proposed LS-DHM with the state-of-the-art on the SUN397 database.}
\begin{tabular}{p{3.3cm}||p{1.6cm}<{\centering}||p{1.6cm}<{\centering}}
\hline
\textbf{Method} & \textbf{Publication} & \textbf{Accuracy}($\%$)\\
\hline

Xiao \emph{et al.}\cite{xiao2010sun} & CVPR2010&38.00\\ \hline
FV(SIFT)\cite{sanchez2013image} &  IJCV2013 &43.02\\
FV(SIFT+LCS)\cite{sanchez2013image}& IJCV2013 &47.20\\
FV(SPM+OPM)\cite{xie2014orientational}& CVPR2014 &45.91\\
LCCD+SIFT \cite{Guo2015} &arXiv2015&49.68\\
\hline
DeCAF \cite{donahue2013decaf} & ICML2014& 40.94\\
MOP-CNN \cite{gong2014multi} &ECCV2014 &51.98\\
Koskela \textit{et al.}\cite{koskela2014convolutional} &ACM2014 &54.70\\
DAG-VGGNet19 \cite{Yang2015}&ICCV2015&56.20\\
Ms-DSP (VGGNet16) \cite{Gao2015} &arXiv2015&59.78\\
C-HLSTM \cite{Zuo2015} &arXiv2015&60.34\\
\hline\hline
Places-CNN (AlexNet)\cite{zhou2014learning} &NIPS2014& 54.32\\
LS-DHM (AlexNet) &--&62.97\\
\hline
GoogleNet &--& 58.79\\
LS-DHM (GoogleNet) &--& 65.40\\\hline
VGGNet11 &--& 64.02\\
LS-DHM (VGGNet11) &--& \textbf{67.56}\\\hline

\hline
\end{tabular}
\
\label{table3}
\end{table}

We further investigate the performance of our LS-DHM by using various FC-features. The LS-DHM obtains consistent large improvements over corresponding baselines, regardless of the underlying FC-features, and achieves the state-of-the-art results on both benchmarks. It obtains 83.75\% and 67.56\% accuracies on the MIT Indoor67 and the SUN397 respectively, outperforming the strong baselines of 11-layer VGGNet with about 4\% improvements in both two datasets. On the MIT Indoor67, our results are compared favourable to the closest performance at $81.0\%$ obtained by the FV-CNN \cite{cimpoi2015deep}, which also explores the convolutional features from a larger-scale 19-layer VGGNet. On the SUN397, we gain a large 7\% improvement over the closest result archived by the C-HLSTM \cite{Zuo2015}, which integrates the CNN with hierarchical recurrent neural networks (C-HLSTM). The sizable boost in performance on both benchmarks convincingly confirm the promise of our method. For different FC-features, we note that the LS-DHM obtains larger improvements on the AlexNet and GoogleNet (about 7-8\%), which are about twice of the improvements on the VGGNet. This may due to the utilization of very small 3$\times$3 convolutional filters by the VGGNet. This design essentially captures more local detailed information than the other two. Thus the proposed FCV may compensate less to the VGGNet.

\section{ Conclusions}

We have presented the Locally-Supervised Deep Hybrid Model (LS-DHM) that explores the convolutional features of the CNN for scene recognition. We observe that the FC representation of the CNN is highly abstractive to global layout of the image, but is not discriminative to local fine-scale object cues. We propose the Local Convolutional Supervision (LCS) to enhance the local semantics of fine-scale objects or regions in the convolutional layers. Then we develop an efficient Fisher Convolutional Vector (FCV) that encodes the important local semantics into an orderless mid-level representation, which compensates strongly to the high-level FC-features for scene classification. Both the FCV and FC-features are collaboratively employed in the LS-DHM representation, leading to substantial performance improvements over current state-of-the-art methods on the MIT Indoor67 and SUN 397.

{\small
\bibliographystyle{IEEEtran}
\bibliography{ref}
}

\end{document}